\documentclass{article}

\usepackage{arxiv}

\usepackage[utf8]{inputenc} % allow utf-8 input
\usepackage[T1]{fontenc}    % use 8-bit T1 fonts
\usepackage{hyperref}       % hyperlinks
\usepackage{url}            % simple URL typesetting
\usepackage{booktabs}       % professional-quality tables
\usepackage{amsfonts}       % blackboard math symbols
\usepackage{nicefrac}       % compact symbols for 1/2, etc.
\usepackage{microtype}      % microtypography
\usepackage{lipsum}
\usepackage{algorithm2e}
\usepackage{graphicx}
\graphicspath{ {./images/} }

\title{An efficient deep reinforcement learning environment for flexible job-shop scheduling}

\author{
 Xinquan Wu \\
 College of Computer Science and Technology\\
 Nanjing University of Aeronautics and astronautics\\
 Nanjing, China 210016\\
  \texttt{wuxinquan@nuaa.edu.cn}
   \And
 Xuefeng Yan \\
 College of Computer Science and Technology\\
 Nanjing University of Aeronautics and astronautics\\
 Nanjing, China 210016\\
  \texttt{yxf@nuaa.edu.cn}
  \And
 Mingqiang Wei \\
 College of Computer Science and Technology\\
 Nanjing University of Aeronautics and astronautics\\
 Nanjing, China 210016\\
  \texttt{mqwei@nuaa.edu.cn}
  \And
  Donghai Guan \\
 College of Computer Science and Technology\\
 Nanjing University of Aeronautics and astronautics\\
 Nanjing, China 210016\\
  \texttt{dhguan@nuaa.edu.cn}
}

\begin{document}
\maketitle
\begin{abstract}
The Flexible Job-shop Scheduling Problem (FJSP) is a classical combinatorial optimization problem that has a wide-range of applications in the real world. In order to generate fast and accurate scheduling solutions for FJSP, various deep reinforcement learning (DRL) scheduling methods have been developed. However, these methods are mainly focused on the design of DRL scheduling Agent, overlooking the modeling of DRL environment. This paper presents a simple chronological DRL environment for FJSP based on discrete event simulation and an end-to-end DRL scheduling model is proposed based on the proximal policy optimization (PPO). Furthermore, a short novel state representation of FJSP is proposed based on two state variables in the scheduling environment and a novel comprehensible reward function is designed based on the scheduling area of machines. Experimental results on public benchmark instances show that the performance of simple priority dispatching rules (PDR) is improved in our scheduling environment and our DRL scheduling model obtains competing performance compared with OR-Tools, meta-heuristic, DRL and PDR scheduling methods.
\end{abstract}

\section{Introduction}

The Job shop scheduling problem (JSSP) is a classical NP-hard combinatorial optimization problem. It has been studied for decades and has been applied in a wide-range of areas including semiconductor, machine manufacturing, metallurgy, automobile manufacturing, supply chain and other fields\cite{Xiong2022}. The Flexible Job-shop Scheduling Problem (FJSP) is a generalization of JSSP, which allows each operation to be processed on multiple candidate machines\cite{Dauzere2023}. This makes it a more challenging problem with more complex topology and larger solution space.

Extensive researches have been widely explored to solve the FJSP in recent years, including the exact, heuristic, meta-heuristic, hyper-heuristic and reinforcement learning (RL) methods. Exact approaches such as mixed integer linear programming (MILP) \cite{Roshanaei2013} may suffer from the curse of dimension. So it is intractable to find exact scheduling solutions in a given time limit. Heuristic dispatching rules are widely used in the real world manufacturing due to their simple and fast nature. Simple priority dispatching rules (PDR)\cite{Sels2012}, such as the most work remaining (MWKR) possess the advantages of high flexibility and easy implementation, but the performance of PDR methods is not stable for different optimization objectives. Hyper-heuristic methods\cite{Zhang2023}such as genetic programming-based hyper-heuristic (GPHH), provide a way of automatically designing the dispatching rules\cite{Zhang2021}. However, the GP-evolved rules often has a large number of features in the terminal set, making it difficult to identify promising search areas and determine the optimal features.

Different from PDR, meta-heuristic methods such as Tabu Search\cite{Hurink1994}, Genetic Algorithms\cite{Rooyani2020}, Differential Evolution\cite{Du2019} and Particle Swarm Optimization\cite{A2010}, can produce higher solution quality than PDR due to the introduction of iterative, randomized search strategies. Nevertheless, these methods also have shortcomings such as slow convergence speed, difficulty in obtaining global optimal solutions for large-scale problems and are difficult to apply to dynamic scenarios that need real-time decisions. 

Reinforcement learning (RL) is one of the most important branches of machine learning and attracts the interests of researchers from numerous fields especially for scheduling. The early RL scheduling methods represent the scheduling policies using arrays or tables\cite{Aydin2000}, which are only suitable for small-scale scheduling problems because of the curse of dimension. However, the scheduling tasks in the real world usually have a higher dimensional state space and large action space, which limits the applications of RL. Since deep neural networks have a strong fitting capability to  process high-dimensional data, deep reinforcement learning (DRL) has been used to solve scheduling tasks with large or continuous state space and shows great potential to solve various scheduling problems. In DRL scheduling methods, the scheduling policy is usually designed based on convolutional neural network (CNN) in\cite{Feng2022}, multi-layer perception (MLP) in\cite{Du2022A}\cite{Gui2023}, recurrent neural network (RNN) in \cite{Lang2021}, graph neural network (GNN) in\cite{Song2022}\cite{Lei2023}\cite{Zeng2022A}, attention networks in \cite{Wang2023} and other deep neural networks in \cite{Luo2021Real}. The scheduling agent is often trained by RL algorithms such as deep Q-learning(DQN), proximal policy optimization (PPO), Deep Deterministic Policy Gradient (DDPG). However, the above DRL scheduling methods obtained low accurate solutions and some are even inferior to simple PDR. Even though the recent GMAS model\cite{Jing2022} whose policy is designed based on the Graph Convolutional Network (GCN), obtained SOTA results on standard benchmark instances, it is a decentralized Multi-agent reinforcement learning (MARL) model. Besides, the current DRL scheduling methods focus mainly on the design of scheduling agents, overlooking the modeling of scheduling environment which is of vital importance for improving the performance of the DRL scheduling methods.

There are mainly four modeling methods for FJSP: the mathematical\cite{Roshanaei2013}, Petri Nets (PN)-based\cite{Muic2022}, Disjunctive Graph (DG)-based\cite{Blazewicz2000} and simulation-based modeling\cite{Wu2023}. The mathematical modeling methods usually formulates the FJSP as a MILP and the formulated problem is then be solved by optimization methods such as Genetic Algorithm. PN is a versatile modeling tool that can be used to model scheduling problems as they can model parallel activities, resource sharing and synchronization. Both modeling methods are not suitable for DRL scheduling methods. DG is another alternative to model the FJSP which integrates machine status and the operation dependencies, and provides critical structural information for scheduling decisions. However, the DG fails to incorporate dynamically changing state information of FJSP, requiring extra handcrafted node features. Besides, the allocation order provided by the DG model is not strictly chronological. The simulation-based modeling method is to develop algorithms to simulate the laws of activities in the real world. In the current simulation environment of DRL scheduling methods, the scheduling process is promoted by the events in the candidate queue, ignoring the occurrence temporal order of these events.

In this paper, a chronological DRL scheduling environment for the FJSP is presented based on discrete event simulation algorithm where at each decision step, the accurate state changes of the scheduling process are recorded by state variables and a novel comprehensive reward function is designed based on the scheduling area. Furthermore, an end to end DRL model for FJSP is proposed based on the actor-critic PPO. Specifically, a very short state representation is expressed by two state variables which are derived directly from the necessary variables in the environment simulation algorithm and the action space is constructed using six PDRs from the literature. Besides, in order to reduce the computation time for the RL agent, a simple scheduling policy is designed based on the MLP with only one hidden layer and Softmax function.  

The contributions of this work are listed as follows. (1) A basic simulation algorithm based on the chronological discrete event is designed for the FJSP, providing a general environment for DRL scheduling methods. (2) A simple DRL model for the FJSP is proposed based on the PPO where a very short state representation for FJSP is presented, avoiding handcrafted state features and massive experiments for feature selection and a novel comprehensive reward function is proposed based on the scheduling area. (3) Experimental results show that the performance of single PDR is improved in our environment, even better than some DRL scheduling methods and the proposed DRL scheduling model obtains competing performance compared with OR-Tools, meta-heuristic, DRL and PDR scheduling methods.

The rest of the paper is organized as follows. A chronological discrete event simulation-based scheduling environment for FJSP is introduced in Section II. In Section III, the proposed DRL scheduling model for FJSP are constructed. Section IV demonstrates the detailed experiments on standard benchmarks with different sizes and the results are compared and discussed, while conclusions and future work reside in Section V.

\section{The simulation environment for FJSP}
In this section, we first introduced the basics of flexible job shop scheduling problems. Then we defined the storage structure of FJSP and the rules of state changing when scheduling. Finally, the detailed simulation algorithm for FJSP is presented.
\subsection {Flexible Job-shop Scheduling Problems}
The FJSP differs from JSSP in that it allows an operation to be processed by any machine from a given set and different machines may have different processing times. The problem is to assign each operation to an eligible machine such that the maximal completion time (makespan) of all jobs is minimized. It can be presented as FJ$\|$Cmax by the three-field notations \cite{Dauzere2023}.

\begin{figure}
  \begin{center}
  \includegraphics[width=6.5in]{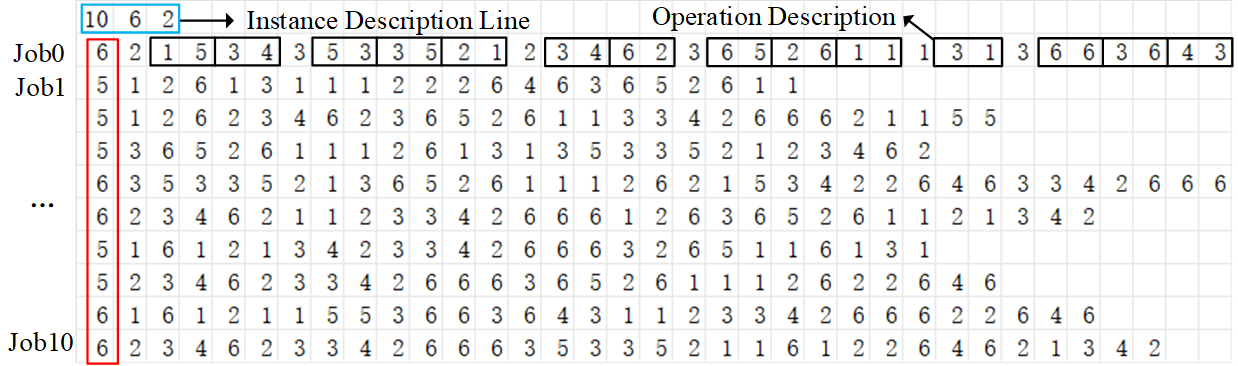}
  \caption{An example of flexible job-shop scheduling problem (MK1\cite{Brandimarte1993}).}\label{Fig instance example}
  \end{center}
\end{figure}

In the public benchmark of FJSP, the environment information is passed through an instance file. As is depicted in Figure \ref{Fig instance example}, the first line in the instance file consists of three integers representing the number of jobs, the number of machines and the average number of machines per operation (optional), respectively. Each of the following lines describes the information of a job where the first integer is the number of operations and the followed integers depicts the operation information. The operation information includes two integers: the first represents the index of a machine and the second is the processing time of the operation on this machine. The processing operation order in a job is advanced from the left to right.
\subsection {The data structure for FJSP}
\begin{figure}
  \begin{center}
  \includegraphics[width=6.5in]{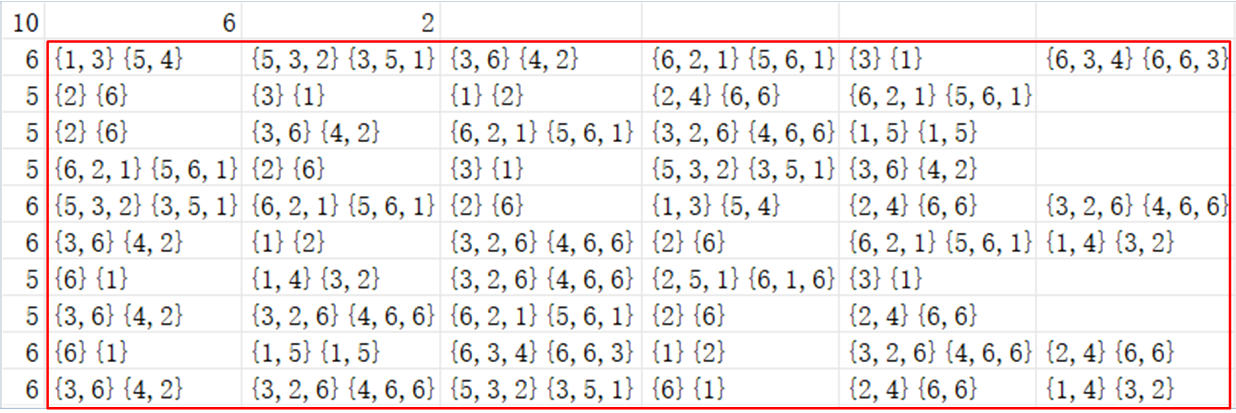}
  \caption{The storage structure of FJSP instance.}\label{Fig storage format}
  \end{center}
\end{figure}

In this paper, based on the benchmark instance file, we proposed a new and efficient storage structure for the FJSP instances. The scheduling information is represented by a two-dimensional table where the job information is described in the rows and the column records the processing stage information. Each element in this table includes two sets: the machine set and the remaining processing time set. Figure \ref{Fig storage format} demonstrates an example of a FJSP instance where the storage structure is marked with a red box. It is efficient to retrieve any operation information using the job index and processing stage index. 
\subsection {Rules of state updating}
\begin{figure}
  \begin{center}
  \includegraphics[width=6.5in]{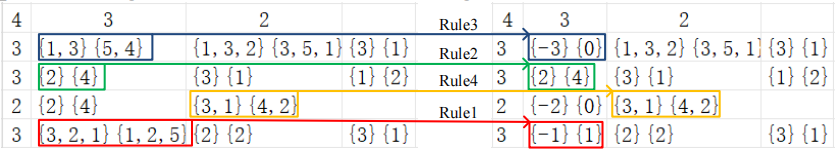}
  \caption{The use of state update rules on a FJSP instance at time 4.}\label{Fig state representation}
  \end{center}
\end{figure}

In order to accurately record the state changes of each scheduling step, four rules of state updating are defined based on the type of operations.  The job operations are divided into four categories: the operations on processing, waiting operations without predecessors, the completed operations and operations whose predecessors have not been completed. The four rules are listed as follows and Figure \ref{Fig state representation} provides an example to demonstrate the use of these rules. 

Rule1: For an operation on processing, the machine set has only one element which is the negative index of the selected machine while the remaining time is recorded in the remaining processing time set and its value decreases as the time advances until the completion of this operation.

Rule2: For a waiting operation without predecessors whose needed machine is occupied, the elements in the machine set are all negative index of machines and the elements in the remaining processing time set stay unchanged. When the needed machine is released, the value of machine index is restored to be positive.

Rule3: For a completed operation, the values in the machine set and remaining processing time set equal to the negative value of machine index and zero, respectively.

Rule4: For operations whose predecessors have not been processed, the values in the machine set and remaining processing time set remain unchanged.

This representation for the FJSP instance can be recorded in a instance file and reloaded to new environment, which means that it can be applied to the non-zero or re-entrant scenarios.
\subsection {Simulation algorithm for FJSP}
In this paper, the proposed scheduling environment model is a simulation model based on chronological discrete events. There is a timer used to record the current time and the triggering time of events are recorded in a state variable. Once the current time reaches the triggering time of any events, this event will occur and the corresponding event response program will be executed. The detailed process is illustrated in Algorithm \ref{alg:environment for fjsp}.

\RestyleAlgo{ruled}
\SetKwComment{Comment}{/* }{ */}
\begin{algorithm}[hbt!]
\caption{The simulation algorithm for FJSP}\label{alg:environment for fjsp}
Initialize variables \textit{assignable\_job}, current time \textit{T}, \textit{completed\_op\_of\_job}, advanced\_time \textit{ta},
\textit{next\_time\_on\_machine},  \textit{job\_on\_machine}, the number of machines \textit{M}\;
\While{not stop}{
    Get an assignable job from a scheduling policy\;
    Choose a machine for the selected job using PDR\;
    Record the completion time of the current job operation in \textit{next\_time\_on\_machine}\;
    Update the status of jobs and machines which is recorded in \textit{assignable\_job} and \textit{job\_on\_machine}\;
    \textit{scheduling\_area} = -\textit{processing\_time} \;
    \While{there is no assignable job}{
        \eIf{\textit{T} $\textless$ min(\textit{next\_time\_on\_machine})}
        {
            \textit{T}=\textit{min(next\_time\_on\_machine)}\;}
        {
            \textit{T}=\textit{second\_min(next\_time\_on\_machine)}\;
        }
        \While{$m \leq M$}{     
           \textit{ta} = \textit{T} - \textit{next\_time\_on\_machine}[$m$]\;
           \If{\textit{ta} $\textgreater$ 0}{
                \textit{next\_time\_on\_machine}[$m$]+=\textit{ta}\;
            \textit{scheduling\_area} -= \textit{ta}\;
            }   
        }
        {
        \While{$m \leq M$}{
            \textit{left\_time} = \textit{T} - \textit{next\_time\_on\_machine}[$m$]\;
            \If{\textit{left\_time} == 0}{
                Release machine $m$ and its job $j$\;
                \textit{completed\_op\_of\_job}[$j$]+=$1$\;
                Update the status of jobs and machines\;
                \If{the machine for $j$ is occupied}{
                    Update the $j$ to be not assignable\;
                }
            }
            
        }        
        }
    }    
}
\end{algorithm}

The proposed algorithm is mainly composed of four parts: the selection of jobs and machines, the state update of jobs and machines, time advance (the third while statement) and machine release (the fourth while statement). In the first part, the decision action is divided into the PDRs for job and machine selection and then the specific job number and machine number are derived from the PDRs. After allocating the jobs and machines, their states are updated using state variables and table dictionaries. For example, the completion time of this job operation (or the release time of machine) is recorded in the \textit{next\_time\_on\_machine} and the job-machine allocation information is recorded in the \textit{job\_on\_machine}. Once a job is allocated on a machine, the state of this job in the \textit{assignable\_job} changes to zero and the states of the jobs which need that machine, are updated. If the candidate machines of a job are all occupied by other jobs this job will become not assignable.

When there are no assignable jobs, the time advance stage is coming. In this part, the current time is first updated based on the values in \textit{next\_time\_on\_machine} (if the current time is smaller than any value in \textit{next\_time\_on\_machine}, take the smallest value, otherwise take the next smallest of them). The length of time step that needs to advance is then calculated based on \textit{next\_time\_on\_machine} and the current time. Finally, the next time of machines whose next time is slower than current time is advanced to current time. The machine release part releases the machines whose release time reaches the current time and updates the status of jobs and machines. When current job operation is completed, the occupied machine becomes idle and the current operation of this job move to next operation. The states of jobs and machines are updated in the \textit{assignable\_job} and \textit{job\_on\_machine}. If all the operation of current job is completed, this job become not assignable and if the machine needed for the next operation is occupied, this job is still not assignable.

\section {A DRL scheduling framework for FJSP}
\begin{figure}
  \begin{center}
  \includegraphics[width=6.5in]{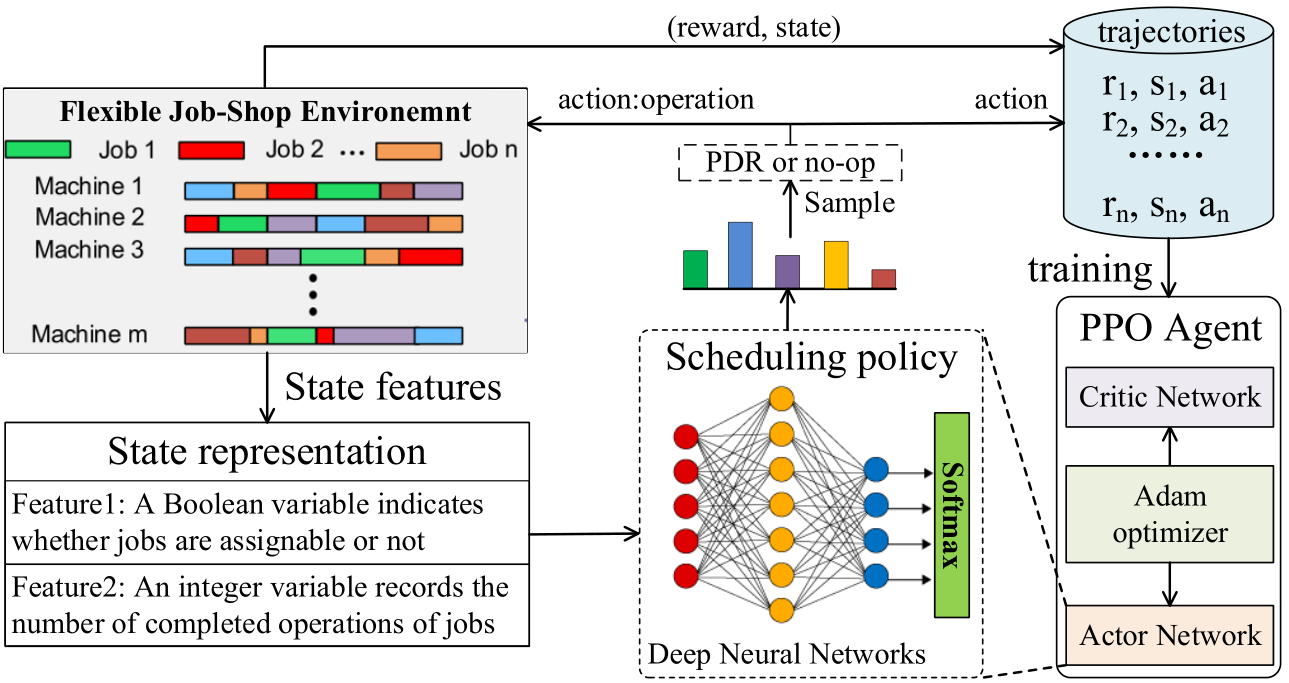}
  \caption{The DRL framework for flexible job-shop scheduling problems.}\label{Fig DRL framework}
  \end{center}
\end{figure}

In this section, we introduced the overall DRL framework for FJSP, which is illustrated in Figure \ref{Fig DRL framework}. It is composed of (1) the the state representation based on two state variables; (2) the PDR action space; (3) the reward function based on scheduling area; (4) the scheduling policy based on deep neural networks and a Softmax function; and (5) the PPO agent with an actor-critic learning architecture. 
\subsection {State representation based on state variables}
State representation depicts the features of scheduling environment and determines the size of the state space, which is very crucial in RL scheduling methods. Various state representations for FJSP are presented in the literature and are mainly focused on the disjunctive graph\cite{Song2022}\cite{Lei2023}, feature matrix\cite{Wu2023} and handcrafted state variables \cite{Du2022A}\cite{Gui2023}\cite{Lang2021}. However, whether it is the node features of the disjunctive graph or the matrix, or variable features, the state features of the job shop are mainly manually designed, which requires a large amount of professional domain knowledge, computing resources and time.

In this paper, a novel short state representation is proposed to reduce feature redundancy and to avoid handcrafted feature design and feature selection, which requires less computation time of environment state variables and scheduling policy networks. Two state variables: \textit{assignable\_job} and \textit{completed\_op\_of\_job} are selected from the Algorithm \ref{alg:environment for fjsp} as the state features of the job shop scheduling environment. The variable \textit{assignable\_job} is a Boolean vector to represent whether a job can be allocated or not. The \textit{completed\_op\_of\_job} represents the number of completed operation of a job and this value is then scaled by the maximum number of operations in the jobs to be in the range [0,1]. The length of both variables equals the number of jobs. Finally, in order to represent the environment state, the scaled variables are simply concatenated to a vector whose length equals two times of the number of jobs.

There are many advantages of our state representation: (1) the length of state features is much less than that in previous research which means less computation time of environment state variables and scheduling policy networks; (2) the state features are unique in a scheduling solution, which suggests that the state features can be easily distinguished by scheduling agent; and (3) the state features are directly derived from the two state variables in Algorithm \ref{alg:environment for fjsp}, avoiding the massive experiments on feature design and selection. 

\subsection {Action space based on PDR}
In DRL scheduling methods based on single agent, the action is the output of scheduling policy networks, which is usually an integer. Since the FJSP needs to select a job and a machine at each decision, we decomposed this integer into two parts by dividing it by the number of PDRs for machine selection where the quotient is the index of PDR for jobs assignment and remainder represents the index of PDR for machine selection. 

In this paper, six PDRs are selected to construct the action space for simplicity of implementation and ease of generalization. For selecting a job, four PDRs are selected directly from the literature \cite{Zhang2020} including the Shortest Processing Time (SPT), Most Work Remaining (MWRK), Most Operations Remaining (MOR) and Minimum ratio of Flow Due Date to Most Work Remaining (FDD/MWRK). The Longest Remaining Machine time not including current operation processing time (LRM) is selected from\cite{Han2020} due to its excellent performance. The First In First Out (FIFO) is selected because it is widely used in various scheduling problems. In addition, the action space for selecting the machine given a job is composed of the SPT and Longest Processing Time(LPT). So that the total size of action space equals the number of PDRs for selecting jobs multiplied by the number of PDRs for selecting machines. The definitions of these PDR are listed as follows:

\begin{itemize}
\item SPT : min $Z_{i,j}$ = $p_{i,j}$
\item LPT : max $Z_{i,j}$ = $p_{i,j}$
\item MWKR: max $Z_{i,j}$ = $\sum_{j}^{n_i} p_{i,j}$
\item FDD/MWKR: min $Z_{i,j}$ = $\sum_{1}^{j} p_{i,j}$/$\sum_{j}^{n_i} p_{i,j}$
\item MOR: max $Z_{i,j}$ = $n_i -j + 1$
\item LRM : max $Z_{i,j}$ = $\sum_{j+1}^{n_i} p_{i,j}$
\item FIFO: max $Z_{i,j}$ = $t - Rt_i$
\end{itemize}
Where $Z_{i,j}$ is the priority index of operation $O_{i,j}$. $p_{i,j}$ is the processing time of operation $O_{i,j}$. $n_i$ is the number of operations for job $J_i$. $j$ is the number of completed operations. $t$ is the current time and $Rt_i$ is the release time of $J_i$. 

\subsection {Reward function based on scheduling area}
In this paper, we proposed a comprehensible reward function based on the scheduling area (Equation \ref{eq reward}) where the processing time of the allocated operations and the time vacancy of all machines is computed after each dispatching action. 

\begin{equation}\label{eq reward}
reward(s_t,a_t) = -p_{a,j} - \sum_{M} vacancy_m(s_t,s_{t+1})
\end{equation}
where $s_t$ is the current state and $s_{t+1}$ is the next state after applying action $a_t$; $a_t$ is the $jth$ operation of a job with the processing time $p_{a,j}$; $vacancy(s_t,s_{t+1})$ is a function returning the total time vacancy on machine set $M$ while transitioning from state $s_t$ to $s_{t+1}$.

\begin{figure}
  \begin{center}
  \includegraphics[width=3.5in]{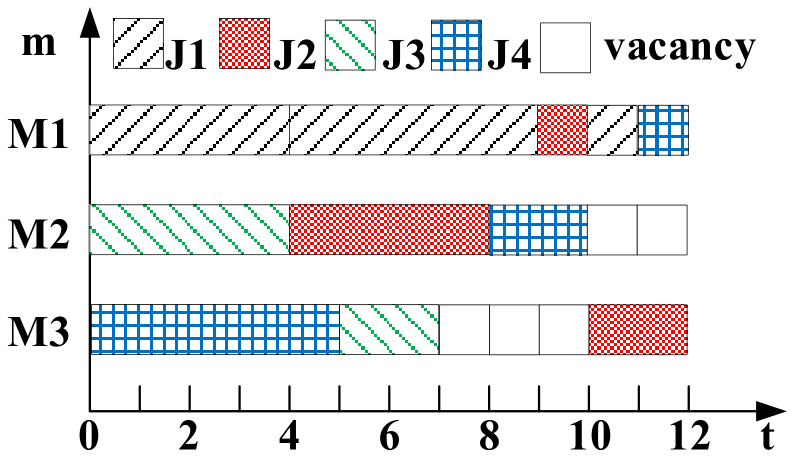}
  \caption{An example to show the scheduling area.}\label{Fig scheduling area}
  \end{center}
\end{figure}

The proposed reward function is motivated by the fact that the total processing time of all jobs and the time vacancy on all machines constructs the scheduling area of all machines which equals the maximum make-span multiplied by the number of machines. As is demonstrated in Figure \ref{Fig scheduling area}, the scheduling area is composed of the total processing time of all jobs (shaded area) and the time vacancy on all machines (white color area). The relationship between accumulated reward and the maximum scheduling make-span is derived from Equation \ref{eq reward make-span}:
\begin{equation}\label{eq reward make-span}
R = -a - b = - (a + b) = - S = - |M| * makespan
\end{equation}
where $R$ is the accumulated reward, $a$ is the total processing time of all jobs, $b$ is the total time vacancy on all machines, $S$ is the scheduling time area and $|M|$is the number of machines.

It is obviously shown from Equation \ref{eq reward make-span} that the total reward and the maximum make-span are negatively linearly dependent and the coefficient is the number of machines. That is, minimizing the maximum scheduling make-span is equivalent to maximizing the total reward.

\subsection {Model training method based on PPO}

In order to strengthen the representation ability of reinforcement learning scheduling method, the scheduling policy is usually represented by deep neural networks such as CNN, RNN and MLP. In our method, the state feature vector is first fed to the MLP to obtain a scalar score for each action and a Softmax function is then applied to output a distribution over the computed score, which is shown in Equation \ref{eq policy}. The structure of our scheduling policy is demonstrated in Figure \ref{Fig DRL framework}, which constructs the actor network of PPO agent. Similar to the actor network, the critic network is implemented by the MLP with one hidden layer. The parameters of our scheduling policy are learned by a clipped PPO whose loss function is expressed in Equation \ref{eq loss}.

\begin{equation}\label{eq policy}
p(a_t|s_t) = Softmax(MLP_{\pi_\theta}(s_t))
\end{equation}
Where $p(a_t|s_t)$ is the selection probability of action $a_t$ at time t in state $s_t$ and $\theta$ is the parameter of the scheduling policy $\pi$.
\begin{equation}\label{eq loss}
L(\theta) = E_t[min(r_t(\theta)A_t, clip(r_t(\theta), 1-\epsilon, 1+ \epsilon)A_t)]
\end{equation}
Where $r_t(\theta)=\frac{\pi_\theta(a_t|s_t)}{\pi_{\theta_{old}}(a_t|s_t)}$; $A_t$ is an estimator of the advantage function at time step $t$; $clip$ is a clipping function and $\epsilon$ is a hyper parameter which is used to limit the boundary of the objective function.

\RestyleAlgo{ruled}
\SetKwComment{Comment}{/* }{ */}
\begin{algorithm}[hbt!]
\caption{Model training method based on PPO }\label{alg:hper ppo for FJSP}
Initialize maximum episode $N$, discount factor $\gamma$, the number of trajectories $T$ buffer $B$, buffer$M$, batch size $b$, update steps $K$; PER number $C$; actor, critic networks and corresponding optimizers\;
\While{$e \leq N$}{
    \While{$t \leq T$}{
        Reset schedule scheme and clear old results\; 
        Observe state $s_0$ \;
        \While{true}{
           Select action $a_t$ and get its probability $p_a$\;
           Execute action $a_t$ in the environment\;
           Observe reward $r_t$ and the next state $s_{t+1}$\;
           Store transition($s_t$, $a_t$, $r_t$, $p_a$) in $B$\;
           \If{done}{
                Calculate discounted reward using $\gamma$\;
                Add $B$ to$ M$ and clear $B$\;
                break\;
            }   
        }
    }{
        \While{$k \leq K$}{
            Divide $M$ randomly into $S$ subsets with size $b$\;
            \While{$s \leq S$}{
                Optimize the actor and critic networks\;
                Calculate the priority of experience samples\;
            }
            {
                \While{$c \leq C$}{
                    Resample $b$ transitions using the priority\;
                    Calculate the importance sampling weights\;
                    Optimize update actor and critic networks\;
            }
            }
        }
        \If{the result is convergent or time out}{
            Save the scheduling networks and results\;
            \textbf{break}\;
        }
    }
    
}
\end{algorithm}

The detailed training process is provided by Algorithm \ref{alg:hper ppo for FJSP}. The training process includes two aspects: data collection and policy learning. When collecting training data, \textit{T} independent complete trajectories of scheduling are generated and they are collected in the memory buffer \textit{M}. On the policy learning stage, the training data are used for \textit{K} times. At each time, these data are randomly divided into batches whose length is related to the scale of instances and the scheduling agent learns from each batch data. After replaying these experience samples evenly, prioritized experience replay is performed for \textit{C} times. The training process will stop until the episode reaches the maximum iterations or the result is convergent or the scheduling is time out. We defined that the result is thought as convergent if the make-span values are the same in 30 decision steps and the training time is limited in an hour.

% === III. Experiments =======================================
% =================================================================================
\section{Experiments}

In this section, in order to show the effectiveness of our DRL scheduling environment for FJSP (Algorithm \ref{alg:environment for fjsp}) and to evaluate the performance of the proposed DRL scheduling methods, a group of experiments are performed on public FJSP benchmarks with various sizes and the results are compared with different type of scheduling methods from the recent literature. Finally, the training details such as the training time are demonstrated. 

\subsection {Benchmark instances and baseline models}

In this paper, two well-known benchmarks of FJSP are used to evaluate our proposed methods including the MK instances (MK01-MK10) in \cite{Brandimarte1993} and the three group of LA instances (edata, rdata and vdata each with 40 instances) in\cite{Hurink1994}. 

This paper compared with PDR, exact solver, meta-heuristic and DRL models. Six PDRs are used to compare the scheduling results where four out of six PDRs are compared in old scheduling environment and our proposed environment. We also compared with the well-known Google \href{https://developers.google.com/optimization}{OR-Tools} which is a powerful constraint programming solver showing strong performance in solving industrial scheduling problems. For meta-heuristic scheduling methods, two recent improved Genetic Algorithms are selected from\cite{Nguyen2017} and \cite{Chen2020A}. For DRL scheduling methods, we compared with competing models in recent three years, including four DRL methods proposed by Feng et al.\cite{Feng2022}, Zeng et al.\cite{Zeng2022A}, Song et al.\cite{Song2022} and Lei et al.\cite{Lei2023} respectively, the GSMA model\cite{Jing2022} which is a MARL scheduling method and obtains the state-of-the-art results, the DANILE model\cite{Wang2023} which is based on the attention mechanism.

\subsection {Model configurations}
The PPO agent adopts the actor-critic architecture where the actor networks represent the scheduling policy and the critic networks calculate the state-value of policy networks. The actor network is implemented by MLP and Softmax function while the critic network is only represented by the MLP. Both networks are optimized by Adam optimizer, use ReLU as activation function and have only one hidden layer with dynamic hidden dimension which equals the length of state features. For each problem size, we train the policy network for at most 8000 iterations, each of which contains 9 independent trajectories (i.e. complete scheduling process of instances) and we use a dynamic batch size which equals two times of the scale (the total number of operations of all jobs) of an instance. For PPO, we set the epochs of updating network to 10 and the clipping parameter epsilon to 0.2. We set the discount factor $\gamma$ to 0.999 and the learning rate are 1e-3 and 3e-3 for actor and critic network respectively. For prioritized experience replay, the parameter $\alpha$ is set to 0.6, the value of $\beta$ anneals from 0.4 to 1, the number of prioritized experience replay C is set to 1, and the number of convergence training steps is 2000.

The hyper parameters of the number of trajectories, batch size and the dimension of MLP hidden layers are all carefully selected and the other parameters are widely used in the literature or follow the default settings in PyTorch. The experiments run on a laptop equipped with Windows 10 64 operating system 8G RAM, Intel Core i7-9750H 2.60GHz CPU. Our code is available at \href{https://github.com/sx1616039/simple2fjsp}{github}.

\subsection {Results Analysis}

In order to evaluate our proposed scheduling environment, we first tested the performance of the six PDRs in our proposed scheduling environment on the MK benchmark instances where the six PDRs are used to select a job while the SPT is used for the selection of machines. The results are shown in Table \ref{tab:PDR results}. The widely-used four PDRs (SPT, MWKR, FIFO and MOR) are performed in our environment and the results are compared with that in old scheduling environment where their best results are selected from literature\cite{Song2022}. The symbol “-” means that the specific results are not recorded in the literature. The best scheduling result of all six PDRs on each instance are recorded in minPDR row. Besides, the results of two DRL scheduling methods proposed respectively by Feng et al.\cite{Feng2022} and Zeng et al.\cite{Zeng2022A} are also compared in Table \ref{tab:PDR results} and their results are directly from their literature instead of our implementation environment.

As is shown in Table \ref{tab:PDR results}, the results indicate that the performance of SPT, FIFO and MOR in our environment is improved while the performance of MWKR is suddenly worse than before. The LRM obtained the best average results among all the six PDRs and event better than some DRL scheduling methods such as the models proposed by Feng et al. and Zeng et al., which is surprisingly interesting. The best result on different instances is distributed in different PDRs, such as the MWKR obtained the best results on MK1 and FIFO performs best on MK3. So that, the minPDR can get better results than LRM.

\begin{table*}
\caption{\label{tab:PDR results}The scheduling results of PDRs on MK benchmark instances}
\centering
\begin{tabular}{c c c c c c c c c c c c}
\hline									 		
Models & MK1 & MK2 & MK3 & MK4 & MK5 & MK6 & MK7 & MK8 & MK9 & MK10 & AVG \\\hline
SPT-old	&	-	&	-	&	-	&	-	&	-	&	-	&	-	&	-	&	-	&	-	&	238.8	\\
SPT	&	55	&	55	&	268	&	78	&	235	&	85	&	227	&	601	&	431	&	329	&	236.4	\\
MWKR-old	&	-	&	-	&	-	&	-	&	-	&	-	&	-	&	-	&	-	&	-	&	200.17	\\
MWKR	&	44	&	39	&	223	&	83	&	189	&	82	&	256	&	523	&	334	&	280	&	205.3	\\
FIFO-old	&	-	&	-	&	-	&	-	&	-	&	-	&	-	&	-	&	-	&	-	&	206.09	\\
FIFO	&	46	&	37	&	211	&	80	&	192	&	85	&	211	&	531	&	347	&	277	&	201.7	\\
MOR-old	&	-	&	-	&	-	&	-	&	-	&	-	&	-	&	-	&	-	&	-	&	202.31	\\
MOR	&	49	&	37	&	234	&	75	&	191	&	79	&	203	&	533	&	347	&	252	&	200	\\
LRM	&	49	&	36	&	216	&	74	&	186	&	87	&	189	&	523	&	325	&	254	&	193.9	\\
FDD/MWKR	&	52	&	40	&	231	&	92	&	200	&	97	&	232	&	552	&	361	&	289	&	214.6	\\
minPDR	&	44	&	36	&	211	&	74	&	186	&	79	&	189	&	523	&	325	&	252	&	191.9	\\
Feng\cite{Feng2022}	&	42	&	32	&	204	&	78	&	187	&	90	&	169	&	531	&	349	&	279	&	196.1	\\
Zeng\cite{Zeng2022A}	&	48	&	34	&	235	&	77	&	192	&	78	&	190	&	544	&	375	&	256	&	202.9	\\
\hline
\end{tabular}
\end{table*}

The second group of experiments are also performed on MK benchmark instances to evaluate the generality of our proposed DRL scheduling agent in our environment. We train these instances independently for 5 times and the average results are recorded in “Ours” column. As is shown in Table \ref{tab:DRL results}, we compared the performance of our proposed DRL scheduling model with the exact solver (the OR-Tools), the meta-heuristic scheduling methods (2SGA\cite{Nguyen2017} and SLGA\cite{Chen2020A}), DRL scheduling methods (GMAS\cite{Jing2022}, DANILE\cite{Wang2023}, and models proposed by Song et al.\cite{Song2022}) and minPDR method. The LB and the UB columns respectively record the lower bound and upper bound scheduling results in the literature. The results in these compared methods are directly from the literature except the results of PDRs which are performed in our environment. Our results are printed in boldface.
\begin{table*}
\caption{\label{tab:DRL results}Make-span performance of our DRL model on MK benchmark instances}
\centering
\begin{tabular}{c c c c c c c c c c c c}
\hline				 		
Models	&	MK1	&	MK2	&	MK3	&	MK4	&	MK5	&	MK6	&	MK7	&	MK8	&	MK9	&	MK10	&	AVG \\\hline
LB	&	36	&	24	&	204	&	48	&	168	&	33	&	133	&	523	&	299	&	165	&	163.3	\\
UB	&	42	&	32	&	211	&	81	&	186	&	86	&	157	&	523	&	369	&	296	&	198.3	\\
GMAS\cite{Jing2022}	&	39	&	26	&	204	&	60	&	172	&	58	&	137	&	523	&	307	&	197	&	172.3	\\
OR-tools	&	-	&	-	&	-	&	-	&	-	&	-	&	-	&	-	&	-	&	-	&	174.9	\\
2SGA\cite{Nguyen2017}	&	-	&	-	&	-	&	-	&	-	&	-	&	-	&	-	&	-	&	-	&	175.2	\\
\textbf{Ours}	&	\textbf{42}	&	\textbf{29.2}	&	\textbf{204}	&	\textbf{69}	&	\textbf{176}	&	\textbf{73.4}	&	\textbf{152.4}	&	\textbf{523}	&	\textbf{311}	&	\textbf{224.2}	&	\textbf{180.4}	\\
DANIEL\cite{Wang2023}	&	-	&	-	&	-	&	-	&	-	&	-	&	-	&	-	&	-	&	-	&	180.8	\\
SLGA\cite{Chen2020A}	&	40	&	27	&	204	&	60	&	172	&	69	&	144	&	523	&	320	&	254	&	181.3	\\
Song\cite{Song2022}	&	-	&	-	&	-	&	-	&	-	&	-	&	-	&	-	&	-	&	-	&	190.4	\\
minPDR	&	44	&	36	&	211	&	74	&	186	&	79	&	189	&	523	&	325	&	252	&	191.9	\\
\hline
\end{tabular}
\end{table*}

The average make-span is usually used to evaluate the accuracy of the scheduling solutions. As demonstrated in Table \ref{tab:DRL results}, the GMAS which is a DRL scheduling method based on MARL, obtained the best average result. Nevertheless, they used the optimal results of each instance rather than the average results. The average make-span of our proposed DRL model is smaller than the SLGA, DANILE, Song, and minPDR methods and is larger than the OR-Tools, 2SGA and GMAS methods, which shows that the performance of DRL scheduling methods are getting close to meta-heuristic and exact solver. However, different from the complex scheduling networks of GMAS, our model is simple and easy to train to converge, which is stable. 

\begin{figure}
  \begin{center}
  \includegraphics[width=6.5in]{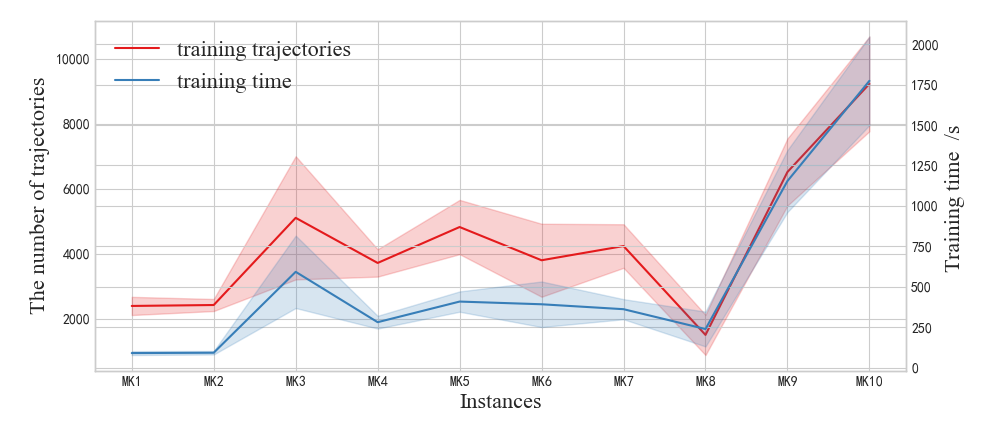}
  \caption{The training trajectories and training time on MK benchmarks.}\label{Fig MK}
  \end{center}
\end{figure}

Beside, we evaluated the convergence performance of our DRL scheduling method. As is demonstrated in Figure \ref{Fig MK}, our DRL model on all the MK instances is convergent in half an hour on average and the number of needed trajectories is less than 900. The convergence time of eight out of ten instances is less than 600 seconds, which is close to the industrial time limitation.

\begin{table}
\caption{\label{tab:LA results}The average make-span comparison on LA instances}
\centering
\begin{tabular}{c c c c}
\hline
Models	&	edata	&	rdata	&	vdata	\\\hline
LB	&	1005.1	&	923.4	&	918.8	\\
UB	&	1028.9	&	934.3	&	920.9	\\
GMAS\cite{Jing2022}	&	1024.7	&	932.2	&	919.6	\\
2SGA\cite{Nguyen2017}	&	-	&	-	&	923.9	\\
OR-Tools	&	1026.7	&	938.4	&	924.4	\\
\textbf{Ours}	&	\textbf{1079.7}	&	\textbf{972.3}	&	\textbf{924.6}	\\
DANIEL\cite{Wang2023}	&	1116.7	&	978.28	&	925.4	\\
Song\cite{Song2022}	&	1116.4	&	986.1	&	931.5	\\
minPDR	&	1155.8	&	1016.5	&	941.6	\\
Lei\cite{Lei2023}	&	-	&	-	&	954.1	\\
\hline
\end{tabular}
\end{table}
In order to show the stability and generality of our proposed DRL scheduling agent in our environment, more experiments are performed on LA benchmark instances, including the edata, rdata and vdata, whose flexibility is getting increased. As is shown in Table \ref{tab:LA results}, we compared with various scheduling methods in the literature such as the exact solver OR-Tools, the meta-heuristic 2SGA, the DRL scheduling methods: the DANILE, GMAS, Lei\cite{Lei2023} and Song as well as the PDR methods. The results of these methods are directly from the literature. The results of PDR methods are from the running in our proposed environment and only the minimum result of six PDRs on each instance is recorded in the minPDR column.

As is demonstrated in Table \ref{tab:LA results}, our proposed DRL scheduling agent got smaller average make-span than the DANILE, Lei, Song and PDR models and obtained larger average make-span than the OR-Tools, GMAS and 2SGA, which shows the competing performance of our proposed DRL scheduling agent in our environment. More interestingly, the minPDR obtained smaller make-span than Lei model which is used to solve large-scale dynamic FJSP. That shows the efficiency of our proposed environment again.

\begin{figure}
  \begin{center}
  \includegraphics[width=6.5in]{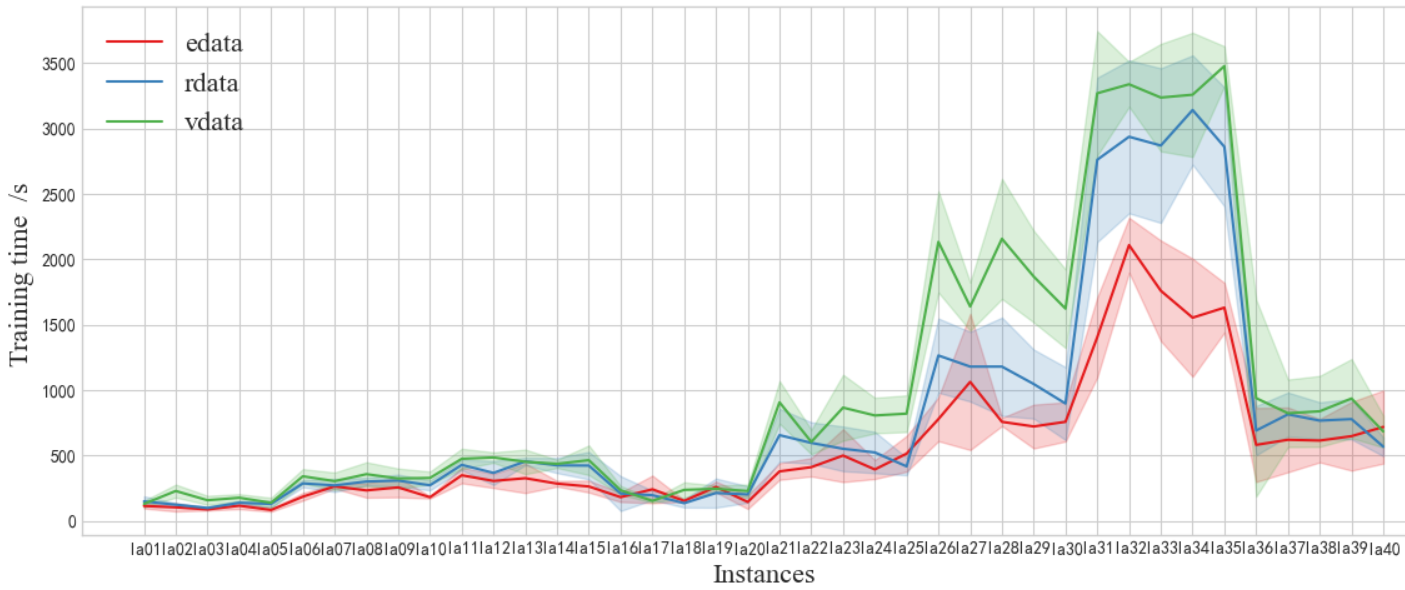}
  \caption{The training time on edata, rdata and vdata instances.}\label{Fig LA}
  \end{center}
\end{figure}

Finally, the training time of our proposed DRL scheduling agent on edata, rdata and vdata are depicted in Figure \ref{Fig LA}. Our proposed DRL model can be trained to converge in an hour on all the benchmark instances and the total training time on edata, rdata and vdata are 22064, 31650 and 40137 seconds, respectively, which shows the training time rises as the increase of the complicity of scheduling instances. 

\section{Conclusion}
In this paper, we proposed a chronological discrete event simulation based DRL environment for FJSP. The scheduling process is described by a simulation algorithm where the state changes are captured by state variables and the reward function is calculated based on the scheduling area of machines at each decision step. In this simulation environment, an end-to-end DRL framework is proposed based on the actor-critic PPO, providing a flexible infrastructure for the design of state representation, action space and scheduling policy networks.  Besides, a simple DRL model for the FJSP is presented by defining the state representation of very short state features based on two state variables in the simulation environment, the action space composed of the widely-used priority dispatching rules in the literature and the scheduling policy based on the MLP with only one hidden layer. Various experiments are performed on classic benchmark instances with different sizes and the results show that the performance of PDR is improved in our environment, even better than some DRL methods. Besides, our DRL scheduling method provides competing scheduling performance compared with the DRL, meta-heuristic and PDR methods. 

Future research will mainly focus on the design of scheduling policy networks as well as the state representation. The state features can be thought as texts or images so that various networks in the NLP and CV fields can be used in the scheduling policy networks, such as SPP networks, TextCNN and Transformer.

\bibliographystyle{plain}
\bibliography{references} 
\end{document}